\pdfoutput=1
\documentclass[preprint,12pt]{elsarticle}

\usepackage{acro}
\usepackage{amsfonts}
\usepackage{amsmath}
\usepackage{amssymb}
\usepackage{amsthm}
\usepackage{bbm}
\usepackage{algorithm}
\usepackage{algpseudocode}
\usepackage{url}
\usepackage{subcaption}
\usepackage{booktabs}
\usepackage{colortbl}
\usepackage{multirow}

\DeclareAcronym{Adam}{
      short = Adam,
      long = \textit{adaptive moment estimation}
}

\DeclareAcronym{ANOVA}{
      short = ANOVA,
      long = \textit{analysis of variance}
}

\DeclareAcronym{Adagrad}{
      short = Adagrad,
      long = \textit{adaptive gradients}
}

\DeclareAcronym{Adadelta}{
      short = Adadelta,
      long = \textit{adaptive learning rate}
}

\DeclareAcronym{RMSProp}{
      short = RMSProp,
      long = \textit{root mean squared error propagation}
}

\DeclareAcronym{PSO}{
      short = PSO,
      short-plural-form = PSOs,
      long = \textit{particle swarm optimisation},
      long-plural-form = \textit{particle swarm optimisers}
}

\DeclareAcronym{Momentum}{
      short = Momentum,
      long = \textit{momentum}
}

\DeclareAcronym{NAG}{
      short = NAG,
      long = \textit{Nesterov accelerated gradients}
}

\DeclareAcronym{GA}{
      short = GA,
      short-plural-form = GAs,
      long = \textit{genetic algorithm},
      long-plural-form = \textit{genetic algorithms}
}

\DeclareAcronym{GP}{
      short = GP,
      long = \textit{genetic programming},
}

\DeclareAcronym{AI}{
      short = AI,
      long = \textit{artificial intelligence}
}

\DeclareAcronym{ANN}{
      short = ANN,
      short-plural-form = ANNs,
      long = \textit{artificial neural network},
      long-plural-form = \textit{artificial neural networks}
}

\DeclareAcronym{AN}{
      short = AN,
      short-plural-form = ANs,
      long = \textit{artificial neuron},
      long-plural-form = \textit{artificial neurons}
}

\DeclareAcronym{BHH}{
      short = BHH,
      short-plural-form = BHHs,
      long = \textit{Bayesian hyper-heuristic},
      long-plural-form = \textit{Bayesian hyper-heuristics}
}

\DeclareAcronym{BinXE}{
      short = BinXE,
      long = \textit{binary cross entropy}
}

\DeclareAcronym{CatXE}{
      short = CatEX,
      long = \textit{categorical cross entropy}
}

\DeclareAcronym{FFNN}{
      short = FFNN,
      short-plural-form = FFNNs,
      long = \textit{feedforward neural network},
      long-plural-form = \textit{feedforward neural networks}
}

\DeclareAcronym{HH}{
      short = HH,
      short-plural-form = HHs,
      long = \textit{hyper-heuristic},
      long-plural-form = \textit{hyper-heuristics}
}

\DeclareAcronym{DE}{
      short = DE,
      long = \textit{differential evolution}
}

\DeclareAcronym{GD}{
      short = GD,
      long = \textit{gradient descent}
}

\DeclareAcronym{LReLU}{
      short = LReLU,
      long = \textit{leaky rectified linear unit}
}

\DeclareAcronym{MAP}{
      short = MAP,
      long = \textit{maximum a posteriori estimation}
}

\DeclareAcronym{MLE}{
      short = MLE,
      long = \textit{maximum likelihood estimation}
}

\DeclareAcronym{ML}{
      short = ML,
      short-plural-form = MLs,
      long = \textit{machine learning},
      long-plural-form = \textit{machine learnings}
}

\DeclareAcronym{MAE}{
      short = MAE,
      long = \textit{mean absolute error}
}

\DeclareAcronym{MSE}{
      short = MSE,
      long = \textit{mean squared error}
}

\DeclareAcronym{ReLU}{
      short = ReLU,
      long = \textit{rectified linear unit}
}

\DeclareAcronym{RMSE}{
      short = RMSE,
      long = \textit{root mean squared error}
}

\DeclareAcronym{SparseCatXE}{
      short = SparseCatXE,
      long = \textit{sparse categorical cross entropy}
}

\DeclareAcronym{SGD}{
      short = SGD,
      long = \textit{stochastic gradient descent}
}

\DeclareAcronym{HMM}{
      short = HMM,
      short-plural-form = HMMs,
      long = \textit{hidden Markov model},
      long-plural-form = \textit{Hidden Markov Models}
}

\DeclareAcronym{BP}{
      short = BP,
      long = \textit{backpropagation}
}

\DeclareAcronym{MH}{
      short = MH,
      short-plural-form = MHs,
      long = \textit{meta-heuristic},
      long-plural-form = \textit{meta-heuristics}
}

\DeclareAcronym{PMF}{
      short = PMF,
      short-plural-form = PMFs,
      long = \textit{probability mass function},
      long-plural-form = \textit{probability mass functions}
}

\DeclareAcronym{DNN}{
      short = DNN,
      long = \textit{deep neural network}
}

\DeclareAcronym{AMALGAM}{
      short = AMALGAM,
      long = \textit{multialgorithm, genetically adaptive multiobjective}
}

\DeclareAcronym{HHBOA}{
      short = HHBOA,
      long = \textit{hyper-heuristic Bayesian optimisation algorithm}
}

\DeclareAcronym{BOHTA}{
      short = BOHTA,
      long = \textit{bi-objective hyperheuristic training algorithm}
}

\journal{Information Sciences}

\begin{document}

\begin{frontmatter}
	\title{\tnoteref{note1}Training Feedforward Neural Networks with Bayesian Hyper-Heuristics}

	\tnotetext[note1]{It is recommended that the article be viewed/printed in colour.}




	\author[aff:tuks]{\corref{cor1}A.N.~Schreuder}
	\ead{an.schreuder@up.ac.za}
	\author[aff:tuks]{A.S.~Bosman}
	\ead{annar@cs.up.ac.za}
	\author[aff:stellenbosch]{A.P.~Engelbrecht}
	\ead{engel@sun.ac.za}
	\author[aff:wits]{C.W.~Cleghorn}
	\ead{christopher.cleghorn@wits.ac.za}

	\cortext[cor1]{Corresponding author}

	\begin{abstract}
		The process of training \acfp{FFNN} can benefit from an automated process where the best heuristic to train the network is sought out automatically by means of a high-level probabilistic-based heuristic. This research introduces a novel population-based \acf{BHH} that is used to train \acfp{FFNN}. The performance of the \acs{BHH} is compared to that of ten popular low-level heuristics, each with different search behaviours. The chosen heuristic pool consists of classic gradient-based heuristics as well as \acfp{MH}. The empirical process is executed on fourteen datasets consisting of classification and regression problems with varying characteristics. The \acs{BHH} is shown to be able to train \acp{FFNN} well and provide an automated method for finding the best heuristic to train the \acp{FFNN} at various stages of the training process.
	\end{abstract}






\end{frontmatter}

\section{Introduction}
\label{sec:introduction}

A popular field of focus for studying \acfp{ANN} is the process by which these models are trained. \acp{ANN} are trained by optimisation algorithms known as heuristics. Many different heuristics have been developed and used to train \acp{ANN} \citep{ref:rakitianskaia:2012}. Each of these heuristics has different search behaviours, characteristics, strengths and weaknesses. It is necessary to find the best heuristic to train \acp{ANN} in order to yield optimal results. This process is often non-trivial and time-consuming. Selection of the best heuristic to train \acp{ANN} is often problem specific \citep{ref:allen:1996}.

A recent suggestion related to the field of \textit{meta-learning} is to dynamically select and/or adjust the heuristic used throughout the training process. This approach focuses on the hybridisation of learning paradigms. One such form of hybridisation of learning paradigms is that of hybridisation of different \textit{heuristics} as they are applied to some optimisation problem \citep{ref:burke:2013}. These methods are referred to as \acfp{HH} and focus on finding the best heuristic in \textit{heuristic space} to solve a specific problem.

In the general context of optimisation, many different types of \acp{HH} have been implemented and applied to many different problems \citep{ref:burke:2013}. However, research on the application of \acp{HH} in the context of \acs{ANN} training is scarce. \citet{ref:nel:2021} provides some of the first research in this field, applying a \acs{HH} to \acf{FFNN} training.

This research takes a particular interest in developing a population-based, selection \acs{HH} that makes use of probability theory and Bayesian statistical concepts to guide the heuristic selection process. This paper presents a novel \acf{BHH}, a new high-level heuristic that utilises a statistical approach, referred to as \textit{Bayesian analysis}, which combines prior information with new evidence to the parameters of a selection probability distribution. This selection probability distribution is the mechanism by which the \acs{HH} selects appropriate heuristics to train \acp{FFNN} during the training process.

The selection mechanism implemented by the \acs{BHH} is different from the \acf{AMALGAM} and \acf{BOHTA} methods used by \citet{ref:nel:2021}, as well as the \acf{HHBOA} proposed by \citet{ref:oliva:2019}. The key differences include that the \acs{BHH} does not follow an evolutionary approach to the selected low-level heuristics. As such the population does not generate offspring, but rather reuses entities in the population. Furthermore, the \acs{BHH} implements a discrete credit assignment mechanism, not making use of pareto fronts as in the \acs{AMALGAM} and \acs{BOHTA} methods.

The remainder of this article is structured as follows: Section \ref{sec:anns} provides background information on \acp{ANN}. Section \ref{sec:heuristics} provides details on various types of heuristics that have been used to train \acp{FFNN}. Section \ref{sec:hhs} presents background information on \acp{HH} and meta-learning. Section \ref{sec:probability} presents background information on probability theory. Section \ref{sec:bhh} presents the developed \acs{BHH}. Section \ref{sec:methodology} presents a detailed description of the empirical process and the setup of each experiment. Section \ref{sec:results} provides and discusses the results of the empirical study. Section \ref{sec:conclusion} summarises the research that is done along with a brief overview of the findings.

\section{Artifical Neural Networks}
\label{sec:anns}

This research focuses on a particular type of \acs{ANN}, referred to as \acfp{FFNN}. \acp{FFNN} were the first and simplest type of \acp{ANN} developed \citep{ref:schmidhuber:2015} and implement an architecture consisting of input, hidden and output layers by arranging them in sequential order. Furthermore, \acp{FFNN} implement fully connected topologies, where each \acf{AN} in one layer is connected to all the \acp{AN} in the next, without any cycles \citep{ref:zell:1994}. In \acp{FFNN}, information moves forward, in one direction, from the input nodes, through the hidden nodes and finally to the output nodes.

\textit{Training} is the process whereby the weights of the \acs{FFNN} are systematically changed with the aim of improving the \textit{performance} of the \acs{FFNN}. Finding the optimal weights that produce the best performance on a given task is an optimisation problem. The optimisation algorithm used to find the optimal weights is referred to as a \textit{heuristic}. Heuristics search for possible solutions in the solution-space and make use of information from the search space to guide to process.

During the training process, the \acs{FFNN} is exposed to data while trying to produce some target outcome. The degree to which the produced outcome differs from the target outcome is referred to as \textit{loss}. Since training of \acp{FFNN} is an optimisation problem, the goal of the training process is to minimise the loss. The loss is calculated using an error function.

\section{Heuristics}
\label{sec:heuristics}

A heuristic refers to an algorithmic search technique that serves as a guide to a search process where good solutions to an optimisation problem are being sought out. Many different techniques have been used to train \acp{FFNN} \citep{ref:kingma:2014}. At the time of writing, the majority of work that is published on the training of \acp{FFNN} involves the use of gradient-based techniques \citep{ref:nel:2021}.

Gradient-based heuristics are optimisation techniques that make use of derivatives obtained from evaluating the \acs{ANN} error function. In the context of supervised learning, loss functions produce a scalar value that represents the error between the output of the \acs{ANN} and the desired output. When using \acf{GD} to train \acp{ANN}, the gradients of the loss function is used to adjust the weights of the \acs{ANN} in order to minimise the error \citep{ref:engelbrecht:2007}.

There are many variants of gradient-based heuristics. However, they all fundamentally apply the same generic \acs{GD} framework that propagates the error signal backwards through the \acs{ANN}. This algorithm is known as \acf{BP}, was popularised by \citet{ref:werbos:1994}.

The simplest type of \acs{GD} algorithm is referred to as \acf{SGD}, which implements a gradient-based weight update step for each training pattern. In the context of this research, the implementation of \acs{SGD} refers to the mini-batch training implementation of \acs{GD}, where a small batch of training patterns are fed to the \acs{FFNN} at once and the error function is aggregated across all training patterns.

Alternative variants have been proposed that lead to better control over the convergence characteristics of \acs{SGD}. This research focuses on a number of these variants that include \acf{Momentum} \citep{ref:qian:1999}, \acf{NAG} \citep{ref:sutskever:2013}, \acf{Adagrad} \citep{ref:duchi:2011}, \acs{Adadelta} \citep{ref:zeiler:2012}, \acf{RMSProp} \citep{ref:hinton:2012} and \acf{Adam} \citep{ref:kingma:2014}.

Gradient-based heuristics are sensitive to the problem that they are applied to, with hyper-parameter selection often dominating the research focus \citep{ref:bengio:2000}. \citet{ref:blum:2003} mention that since the 1980s, a new kind of approximate algorithm has emerged which tries to combine basic heuristic methods in higher level frameworks aimed at efficiently and effectively exploring a search space. These methods are referred to as \acp{MH}.

The biggest difference between \acp{MH} and gradient-based heuristics is that \acp{MH} make use of meta-information obtained as a result of evaluating the \acs{FFNN} during training and is not limited to information about the search space \citep{ref:blum:2003}. This also means that \acp{MH} do not necessarily require the error function to be differentiable. \citet{ref:blum:2003} provides advantages of \acp{MH} that include the following: they are easy to implement, they are problem independent and do not require problem-specific knowledge, and they are generally designed to find global optima, while gradient-based approaches can get stuck in local optima more often. Similar to gradient-based heuristics, a number of different meta-heuristics have been used to successfully train \acp{FFNN} \citep{ref:rakitianskaia:2012, ref:vanwyk:2014, ref:gupta:1999}. This research takes a particular interest in population-based \acp{MH} that have been used to train \acp{FFNN}. These include \acf{PSO} \citep{ref:shi:1998}, \acf{DE} \citep{ref:price:2006} and \acfp{GA} \citep{ref:fraser:1957}.

\section{Hyper-Heuristics}
\label{sec:hhs}

\citet{ref:burke:2010} define \acp{HH} as search methods or learning mechanism for selecting or generating heuristics to solve computational search problems. \citet{ref:burke:2003} mention that a \acs{HH} is a high-level heuristic approach that, given a particular problem instance and a number of low-level heuristics, can select and apply an appropriate low-level heuristic at each decision point. \acp{HH} implement a form of \textit{meta-learning} that is concerned with the selection of the best heuristic from a pool of heuristics to solve a given problem. It can be said that \acp{HH} are concerned with finding the best heuristic in \textit{heuristic space}, while the underlying low-level heuristics find solutions in the feasible \textit{search/solution space}.

\citet{ref:burke:2010} proposed a classification scheme used to classify \acp{HH}. According to the proposed classification scheme, \acp{HH} are classified in two dimensions. These include the \textit{source of feedback} used during learning and the nature of the \textit{heuristic search space}. For the dimension that involves the source of feedback, \acp{HH} can be classified as either \textit{no learning}, \textit{online learning} or \textit{offline learning}. For the dimension that involves the nature of the \textit{heuristic search space}, \acp{HH} can be classified as either \textit{heuristic selection} or \textit{heuristic generation}. Further distinction is made between \textit{construction} of heuristics and \textit{perturbation} of heuristics.

In the general context of optimisation, many different types of \acp{HH} have been implemented and applied to many different problems. Some notable examples include \citep{ref:burke:2010, ref:dowsland:2007, ref:grobler:2012, ref:vanderstockt:2018}. Research on the application of \acp{HH} in the context of \acs{FFNN} training is still scarce. \citet{ref:nel:2021} provides some of the first research in this field, applying \acs{BOHTA}, a novel adaptation of an evolutionary-based \acs{HH}, known as the \acs{AMALGAM} \acs{HH} \citep{ref:vrugt:2007}, to \acs{FFNN} training. Furthermore, \citet{ref:oliva:2019} provide \acs{HHBOA}, the first use of Bayesian optimisation in a \acs{HH} context. The method proposed by \citet{ref:oliva:2019} uses a Bayesian selection operator to evolve combinations of low-level heuristics  while looking for good problem solutions to a benchmark of optimisation functions, but does not apply a \acs{HH} to the training of \acp{FFNN}.

This research takes a particular interest in a population-based, selection approach for \acp{HH}, with the particular intent of training \acp{FFNN}. In the context of population-based \acp{HH}, an entity pool exists that represents a pool of candidate solutions to the given problem. Each entity in the entity pool is assigned its own low-level heuristic from the heuristic pool. The selection of the best heuristic to apply to a candidate solution is based on the performance of the heuristic relative to that particular candidate solution at a particular point in the search process. Selection methods often make use of probabilistic approaches.

\section{Probability}
\label{sec:probability}

In general, there are two main views to probability and statistics. These include the \textit{frequentist} and the \textit{Bayesian} view of statistics. Naturally, Bayesian statistics is based on Bayes' theorem \citep{ref:bayes:1763}.

Bayesian statistics describe the probability of an event in terms of some belief, based on previous knowledge of the event and the conditions under which the event happened \citep{ref:hackenberger:2019}. Bayes' theorem expresses how a degree of belief, expressed as a probability, should rationally change to account for the availability of related evidence.

One of the many applications of Bayes' theorem is to do statistical inference. Like \acp{FFNN}, Bayesian models need to be \textit{trained}, a process known as \textit{Bayesian analysis}. Bayesian analysis is the process by which prior beliefs are updated as a result of observing new data/evidence.

Bayesian analysis utilises the concept of conjugate priors. \citet{ref:wackerly:2014} state that conjugate priors are prior probability distributions that result in posterior distributions that are of the same functional form as the prior, but with different parameter values. The conjugate prior to a Bernoulli probability distribution is the Beta probability distribution and the conjugate prior to a categorical and multinomial probability distribution is the Dirichlet probability distribution \citep{ref:wackerly:2014}.

\section{Bayesian Hyper-Heuristics}
\label{sec:bhh}

This section presents the novel \acs{BHH}. The general concept of the \acs{BHH} is summarised as follows: the \acs{BHH} implements a high-level heuristic selection mechanism that learns to select the best heuristic from a pool of low-level heuristics. These low-level heuristics are applied to a population of entities, each implementing a candidate solution to a \acs{FFNN}. The  intent of the \acs{BHH} is to optimise both the underlying \acs{FFNN} and the \acs{FFNN} training process. The \acs{BHH} does so by learning the probability that a given heuristic will perform well at a given stage in the \acs{FFNN} training process. These probabilities are then used as heuristic selection probabilities in the next step of the training process.

\begin{figure}[htbp]
	\centering
	\includegraphics[width=1.0\textwidth]{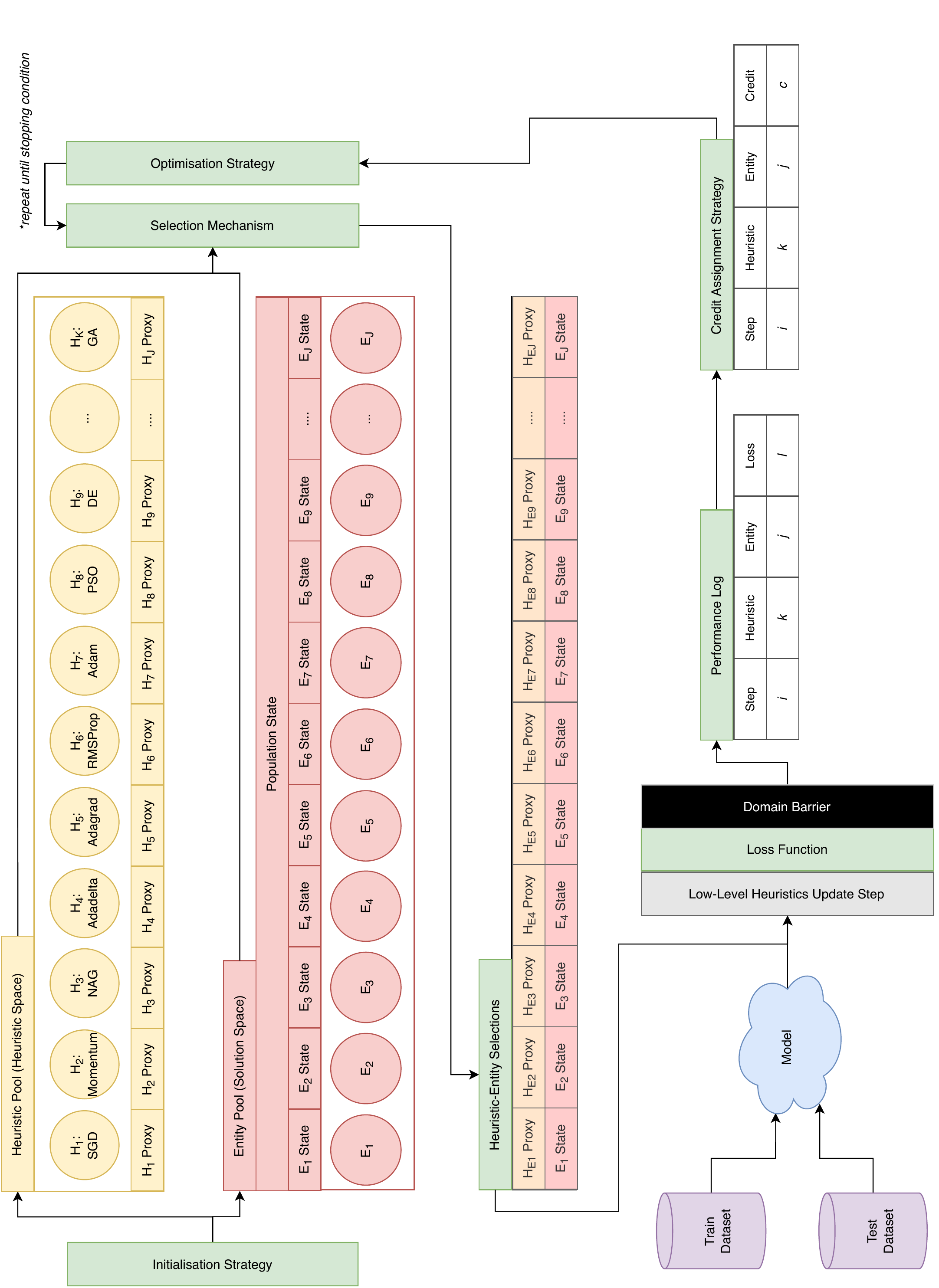}
	\caption[An illustration of
		the architecture and high level components of the \acf{BHH}.]{An illustration of
		the architecture and high level components of the \acf{BHH}.}
	\label{fig:bhh_architecture}
\end{figure}

According to the classification scheme for \acp{HH} by \citet{ref:burke:2010}, the \acs{BHH} is a population-based, meta-hyper-heuristic that utilises selection and perturbation of low-level heuristics in an online learning fashion. Figure \ref{fig:bhh_architecture} provides an illustration of the high-level architecture of the \acs{BHH}. Algorithm \ref{algo:bhh} provides the high level pseudo-code implementation of the \acs{BHH}. Discussions follow on the most important components of the \acs{BHH}.

\begin{algorithm}[htbp]
	\footnotesize
	\caption{The pseudo-code for the implementation of the \acf{BHH}}
	\label{algo:bhh}
	\begin{algorithmic}
		\State step $\gets 0$

		\State select initial heuristics
		\State initialise population and entities
		\State evaluate entities' initial position
		\State update population state

		\While{stopping condition not met}
		\For{all entities in entity pool}
		\If{selected heuristic is gradient-based}
		\State get gradients
		\EndIf

		\State apply low-level heuristic and proxy operations
		\State update population state
		\State log performance metrics to performance log

		\If {step < burn-in window size}
		\State select heuristic
		\Else
		\If {step $\mathbin{\%}$ reanalysis interval = 0}
		\State apply Bayesian analysis
		\EndIf

		\If {step $\mathbin{\%}$ reselection interval = 0}
		\State select heuristic
		\EndIf

		\If {step $<$ replay window size}
		\State prune performance log
		\EndIf
		\EndIf
		\EndFor
		\State step $\gets$ step + $1$
		\EndWhile
	\end{algorithmic}
\end{algorithm}

\subsection{Heuristic Pool}\label{sec:bhh:heuristic_pool}

Generally speaking, the heuristic pool is a collection of low-level heuristics under consideration by the \acs{BHH}. The heuristic pool contains the set of low-level heuristics that, together with their performance information, make up the \text{heuristic space}. Importantly, the heuristic pool must consist of a diverse set of low-level heuristics with varying capabilities. This research takes an interest in including both gradient-based heuristics as well as \acp{MH} in the heuristic pool. This approach is referred to as a \textit{multi-method} approach.

\subsection{Proxies}\label{sec:bhh:heuristic:proxies}

Heuristics often maintain a set of parameters that are used to control the behaviour of the heuristic. These parameters are refered to as heuristic \textit{state}. The concept of proxies arises from the sparsity of state as maintained by different heuristics. Since heuristics maintain (possibly) different states, there is an uncertainty of state transition when switching between heuristics. A solution to state indifference is to \textit{proxy} heuristic state update operations. State is then maintained in two parts: primary and proxied state. Primary state refers to the state that is originally maintained by a heuristic. Proxied state refers to the state that is not directly maintained by the heuristic, but can be updated by outsourcing the required state update operation to another heuristic. The \acs{BHH} thus incorporates a mapping of proxied state update operations as given in the example in Table \ref{tab:bhh:heuristic_pool:proxy_mapping_example}.

\begin{table}[htbp]
	\centering
	\caption{An example of a mapping of proxied state update operation maintained by the \acs{BHH}.}
	\label{tab:bhh:heuristic_pool:proxy_mapping_example}%
	\par\bigskip
	\resizebox{0.38\textwidth}{!}{
		\begin{tabular}{ccccc}
			                                                &   & \multicolumn{3}{c}{State Parameter}                                                                                                                                                                                      \\
			\cmidrule{3-5}                                  &   & 1                                                                      & 2                                                                      & 3                                                                      \\
			\cmidrule{3-5}    \multirow{3}[1]{*}{Heuristic} & A & \cellcolor[rgb]{ .776,  .937,  .808}\textcolor[rgb]{ 0,  .38,  0}{n/a} & \cellcolor[rgb]{ 1,  .922,  .612}\textcolor[rgb]{ .612,  .341,  0}{B}  & \cellcolor[rgb]{ .776,  .937,  .808}\textcolor[rgb]{ 0,  .38,  0}{n/a} \\
			                                                & B & \cellcolor[rgb]{ .776,  .937,  .808}\textcolor[rgb]{ 0,  .38,  0}{n/a} & \cellcolor[rgb]{ .776,  .937,  .808}\textcolor[rgb]{ 0,  .38,  0}{n/a} & \cellcolor[rgb]{ 1,  .922,  .612}\textcolor[rgb]{ .612,  .341,  0}{A}  \\
			                                                & C & \cellcolor[rgb]{ .776,  .937,  .808}\textcolor[rgb]{ 0,  .38,  0}{n/a} & \cellcolor[rgb]{ 1,  .922,  .612}\textcolor[rgb]{ .612,  .341,  0}{B}  & \cellcolor[rgb]{ 1,  .922,  .612}\textcolor[rgb]{ .612,  .341,  0}{A}  \\
		\end{tabular}%
	}
\end{table}%
\noindent
From the example given in Table \ref{tab:bhh:heuristic_pool:proxy_mapping_example}, when \index{heuristic}heuristic A is selected, it will outsource state update operations from \index{heuristic}heuristic B for state parameter 2. Heuristic B will outsource from \index{heuristic}heuristic A for state parameter 3. Finally, \index{heuristic}heuristic C will outsource from \index{heuristic}heuristic A and B for state parameters 2 and 3 respectively. In this way, all \index{heuristic}heuristics maintain all the state parameters.

\subsection{Entity Pool}\label{sec:bhh:entity_pool}

The entity pool refers to a collection or \textit{population} of \textit{entities} that each represent a candidate solution to the underlying \acs{FFNN} being trained. The \acs{BHH} selects from the heuristic pool a low-level heuristic to be applied to an individual entity. The outcome of this selection process is a mapping table that tracks which heuristic has been selected for which entity. These heuristic-entity combinations are applied to the underlying \acs{FFNN}. The \acs{BHH} tracks the performance of each of these combinations throughout the training process in a performance log.

Entities represent candidate solutions to the model's trainable parameters (weights) and other heuristic-specific state parameters. These state parameters are referred to as \textit{local} state. Entities are treated as physical \textit{particles} in a hyper-dimensional search environment. Entities model concepts from physics. For example, the candidate solution is represented as the entity's position. The velocity and acceleration is then analogous to the gradient and momentum of the entity respectfully \cite{ref:eberhart:1995}. Examples of entity state parameters, as derived from various low-level heuristics, include entity position, velocity, gradient, position delta, first and second moments of the gradient, the loss, personal best positions, losses, and so on. The entity state parameters are updated by the associated heuristic.

The population state refers to a collection of parameters that are shared between the entities in the population. Population state is also referred to as \textit{global} state and represents the population's memory. The population state generally contains state parameters that are of importance to multiple heuristics, and usually tracks the state of the population and not individual heuristic. Some examples of population state that can arise from different heuristics include the population of entities themselves, the global best entity found so far, the overall best loss achieved thus far, and so on.

\subsection{Performance Log}\label{sec:bhh:performance_log}

Heuristic selection probability is calculated based on heuristic-entity performance over time. Evidence of heuristic-entity performance is thus required for the \acs{BHH} to learn. Historical heuristic-entity performance outcomes are stored in a performance log. The performance log tracks information such as the current step, selected heuristic, associated entity, the loss achieved and so on. Since the performance log can become very big, only a sliding window of the performance history is maintained at each step in the learning process. The sliding window is also referred to as a \textit{replay} window/buffer.

\subsection{Credit Assignment Strategy}
\label{sec:bhh:credit_assignment_strategy}

The credit assignment strategy is a mechanism that assigns a discrete credit indicator to heuristics that perform well, based on their performance metrics such as loss. The credit assignment strategy implements a component of the ``move acceptance'' process as proposed by \citet{ref:ozcan:2006} and addresses the credit assignment problem as discussed by \citet{ref:burke:2010}. A good credit assignment strategy will correctly allocate credit to the appropriate heuristic-entity combination. This research implements the following credit assignment strategies to choose from: \textit{ibest} (iteration best), \textit{pbest} (personal best), \textit{gbest} (global best), \textit{rbest} (replay window best), and \textit{symmetric}, where credit is assigned to all entity-heuristic combinations, regardless of their performance.

\subsection{Selection Mechanism}\label{sec:bhh:selection_mechanism}

The \acs{BHH} implements a probabilistic predictive model based on the fundamentals of the Na\"{\i}ve Bayes algorithm. The \acs{BHH} thus distinguishes between the following events: \textbf{$\boldsymbol{H}$}, the event of observing \textit{heuristics}, \textbf{$\boldsymbol{E}$}, the event of observing \textit{entities}, and \textbf{$\boldsymbol{C}$}, the event of observing \textit{credit assignments} that indicate that the credit assignment \textit{performance criteria} are met. By Bayes' theorem, the selection mechanism implemented by the \acs{BHH} is given as

\begin{equation}
	\label{eq:bhh:selection_mechanism:predictive_model_prop_to}
	\begin{split}
		P(\boldsymbol{H} \vert \boldsymbol{E}, \boldsymbol{C}; \boldsymbol{\theta}, \boldsymbol{\phi}, \boldsymbol{\psi}) &\propto P(\boldsymbol{E} \vert \boldsymbol{H}; \boldsymbol{\phi}) P(\boldsymbol{C} \vert \boldsymbol{H}; \boldsymbol{\psi}) P(\boldsymbol{H} \vert \boldsymbol{\theta})
	\end{split}
\end{equation}

The predictive model thus models the \textit{proportional} probability of the event (selection of) heuristic $\boldsymbol{H}$, given allocation to entity $\boldsymbol{E}$ and credit requirement $\boldsymbol{C}$, parameterised by sampled $\boldsymbol{\theta} \sim Dir(\boldsymbol{\alpha}; K)$, $\boldsymbol{\phi} \sim Dir(\boldsymbol{\beta}; K)^{J}$ and $\boldsymbol{\psi} \sim Beta(\gamma_{1}, \gamma_{0})$. In the aforementioned, $K$ is the heuristic pool size and $J$ is the entity pool size. The parameters $\boldsymbol{\alpha}$, $\boldsymbol{\beta}$, $\gamma_{1}$ and $\gamma_{0}$ are referred to as concentration parameters. The concentration parameters are used to parameterise the prior probability distributions. \ref{app:naive_bayes} provides mathematical derivations of the predictive model.

\subsection{Optimisation Step}\label{sec:bhh:optimisation_step}

The intent of the \acs{BHH} is to gather evidence that can be used to update prior beliefs about which heuristics perform well during training. These beliefs are represented by the concentration parameters $\boldsymbol{\alpha}$, $\boldsymbol{\beta}$, $\gamma_{1}$ and $\gamma_{0}$. A change in prior beliefs is represented by a change in these concentration parameters. Specifically, it can be said that the optimisation process implemented by the \acs{BHH} updates \textit{pseudo counts} of events that are observed in the performance logs. These pseudo counts track the occurrence of a heuristic, an entity, and resulting performance of these two elements. Through the credit assignment strategy, these pseudo counts are biased towards entity-heuristic combinations that meet performance requirements and yield credit allocations.

Generally, there are two different techniques that are used to train Naïve Bayes classifiers. The frequentist approach implements \acf{MLE} and the Bayesian approach implements \acf{MAP}.

\subsubsection{Maximum A Posteriori Estimation}\label{sec:bhh:optimisation_step:map}

\Acs{MAP} is an approach to optimise the values for $\hat{\theta}_{k}$, $\hat{\phi}_{j,k}$ and $\hat{\psi}_{k}$ by optimising the parameters of their probability distributions. This process is referred to as \textit{Bayesian analysis}. Bayesian analysis makes use of the \textit{posterior} probability distribution. The concentration update operations yielded by \acs{MAP}, are given as follows:

\begin{equation}
	\label{eq:bhh:optimisation_step:map:alpha_update_operation}
	\begin{split}
		\alpha_{k}(t+1) = N_{k} + \alpha_{k}(t)
	\end{split}
\end{equation}

\begin{equation}
	\label{eq:bhh:optimisation_step:map:beta_update_operation}
	\begin{split}
		\beta_{j,k}(t+1) = N_{j,k} + \beta_{j,k}(t)
	\end{split}
\end{equation}

\begin{equation}
	\label{eq:bhh:optimisation_step:map:gamma1_update_operation}
	\begin{split}
		\gamma_{1,k}(t+1) = N_{1,k} + \gamma_{1,k}(t)
	\end{split}
\end{equation}

\begin{equation}
	\label{eq:bhh:optimisation_step:map:gamma2_update_operation}
	\begin{split}
		\gamma_{2,k}(t+1) = N_{0,k} + \gamma_{2,k}(t)
	\end{split}
\end{equation}

\noindent
where $N_{k}$ is a summary variable denoting the count of occurrences of heuristic $k$, $N_{j}$ is a summary variable denoting the count of occurrences of entity $j$, $N_{j,k}$ is a summary variable denoting the count of occurrences of heuristic $k$ for entity $j$, $N_{1,k}$ and $N_{0,k}$ are summary variables denoting the count of occurrences where heuristic $k$ meets performance requirements and where heuristic $k$ does not not meet performance requirements.

It can be said that the \acs{BHH} implements a Gaussian process \citep{ref:gortler:2019}. Since the reselection of heuristics happens at regular intervals, the outcome of a selection in one iteration may influence the outcome of another in the next iteration, making the implementation of the \acs{BHH} a \acf{HMM} \citep{ref:rabiner:1986}.

\subsection{Hyper-Parameters}\label{sec:bhh:hyper_parameters}

The following hyper-parameters are implemented by the \acs{BHH}: the \textit{heuristic pool} configures the type of heuristics included in the heuristic pool, the \textit{population size} specifies the number of entities in the entity pool, the \textit{credit assignment strategy} specifies which credit assignment strategy to use, the \textit{reselection interval} determines the frequency of heuristic reselection, the \textit{replay window size} determines the maximum size of the performance log, the \textit{reanalysis interval} determines the frequency at which Bayesian analysis is applied, the \textit{burn in window size} determines the size of an initial window where experience is simply gathered without reanalysis, and finally, the \textit{discounted rewards} and \textit{normalisation} flags toggle scaling modifiers on values assigned by the credit assignment strategies, backwards in the performance log.

\section{Methodology}
\label{sec:methodology}

This section provides the details of the implementation of the empirical process. At a high level the experimental procedure consist of a comparison between the \acs{BHH} and standalone low-level heuristics. A number of datasets, models and heuristics are specified. Throughout the empirical process, a \acs{BHH} baseline configuration is used.

\subsection{BHH Baseline}\label{sec:methodology:baseline_bhh}

The \acs{BHH} baseline is a name given to a specific configuration of the \acs{BHH} which has been found to provide a reasonable baseline performance. The baseline configuration is used as the cornerstone configuration from which all other heuristics and their configurations are evaluated. The \acs{BHH} baseline configuration is given in Table \ref{tab:methodology:bhh_baseline_configuration}. In Table \ref{tab:methodology:bhh_baseline_configuration}, the heuristic pool configuration, \textit{all}, refers to a configuration where the heuristic pool contains all the low-level heuristics, including all gradient-based heuristics and \acp{MH}.

\begin{table}[htb]
	\centering
	\caption{The \acs{BHH} baseline configuration as it is used in the empirical study.}
	\label{tab:methodology:bhh_baseline_configuration}%
	\par\bigskip
	\resizebox{\textwidth}{!}{
		\begin{tabular}{ccccccccc}
			heuristic pool & population & burn in & credit & reselection & replay & reanalysis & normalise & discounted rewards \\
			\midrule
			all            & 5          & 0       & ibest  & 10          & 10     & 10         & false     & false              \\
		\end{tabular}%
	}
\end{table}%

\subsection{BHH vs. Low-Level Heuristics}
\label{sec:methodology:experiments}

For the standalone heuristics experimental group, a number of low-level heuristics are used along with their specified hyper-parameters. Each of these standalone low-level heuristics is compared to that of the \acs{BHH} baseline configuration, across all datasets. The intent of the standalone heuristics experimental group is to determine if the \acs{BHH} baseline configuration can generalise to multiple problems in comparison to individual low-level heuristics.

Additional to the \acs{BHH} baseline configuration, two more \acs{BHH} configurations are included. These include \acs{BHH} configurations where the heuristic pool only makes use of gradient-based heuristics, and a configuration where the heuristic pool only makes use of \acp{MH}. The intent behind the inclusion of these configurations is to determine the effectiveness of multi-method approaches in the heuristic pool applied to training \acp{FFNN}.

\subsection{Heuristics}\label{sec:methodology:heuristics}

Table \ref{tab:methodology:heuristics} contains a list of all the standalone, low-level heuristics that are used as well as their hyper-parameter configurations. In some cases, parameters are changed dynamically throughout the training process using a \textit{decay schedule}. Note from Table \ref{tab:methodology:heuristics} that values that are configured to make use of a decay schedule are presented with the initial value first and the decay rate in brackets next to it.

\begin{table}[htbp]
	\centering
	\caption{Low-level heuristics and their hyper-parameter configurations.}
	\label{tab:methodology:heuristics}%
	\par\bigskip
	\resizebox{0.6\textwidth}{!}{

		\begin{tabular}{llll}
			\textbf{heuristic} & \textbf{configuration}    & \textbf{value} & \textbf{citation}          \\
			\midrule
			sgd                & learning rate             & 0.1 (0.01)     & \citep{ref:sutskever:2013} \\
			momentum           & learning rate             & 0.1 (0.01)     & \citep{ref:sutskever:2013} \\
			                   & momentum                  & 0.9            &                            \\
			nag                & learning rate             & 0.1 (0.01)     & \citep{ref:sutskever:2013} \\
			                   & momentum                  & 0.9            &                            \\
			adagrad            & learning rate             & 0.1 (0.01)     & \citep{ref:duchi:2011}     \\
			                   & epsilon                   & 1E-07          &                            \\
			rmsprop            & learning rate             & 0.1 (0.01)     & \citep{ref:hinton:2012}    \\
			                   & rho                       & 0.95           &                            \\
			                   & epsilon                   & 1E-07          &                            \\
			adadelta           & learning rate             & 1.0 (0.95)     & \citep{ref:zeiler:2012}    \\
			                   & rho                       & 0.95           &                            \\
			                   & epsilon                   & 1E-07          &                            \\
			adam               & learning rate             & 0.1 (0.01)     & \citep{ref:kingma:2014}    \\
			                   & beta1                     & 0.9            &                            \\
			                   & beta2                     & 0.999          &                            \\
			                   & epsilon                   & 1E-07          &                            \\
			pso                & population size           & 10             & \citep{ref:van:2010}       \\
			                   & learning rate             & 1.0 (0.9)      &                            \\
			                   & inertia weight (w)        & 0.729844       &                            \\
			                   & cognitive control (c1)    & 1.49618        &                            \\
			                   & social control (c2)       & 1.49618        &                            \\
			                   & velocity clip min         & -1.0           &                            \\
			                   & velocity clip max         & 1.0            &                            \\
			de                 & population size           & 10             & \citep{ref:mezura:2006}    \\
			                   & selection strategy        & best           &                            \\
			                   & xo strategy               & exp            &                            \\
			                   & recombination probability & 0.9 (0.1)      &                            \\
			                   & beta                      & 2.0 (0.1)      &                            \\
			ga                 & population size           & 10             & \citep{ref:lambora:2019}   \\
			                   & selection strategy        & rand           &                            \\
			                   & xo strategy               & bin            &                            \\
			                   & mutation rate             & 0.2 (0.05)     &                            \\
		\end{tabular}%
	}
\end{table}%

\begin{figure}[htbp]
	\centering
	\includegraphics[width=0.8\textwidth]{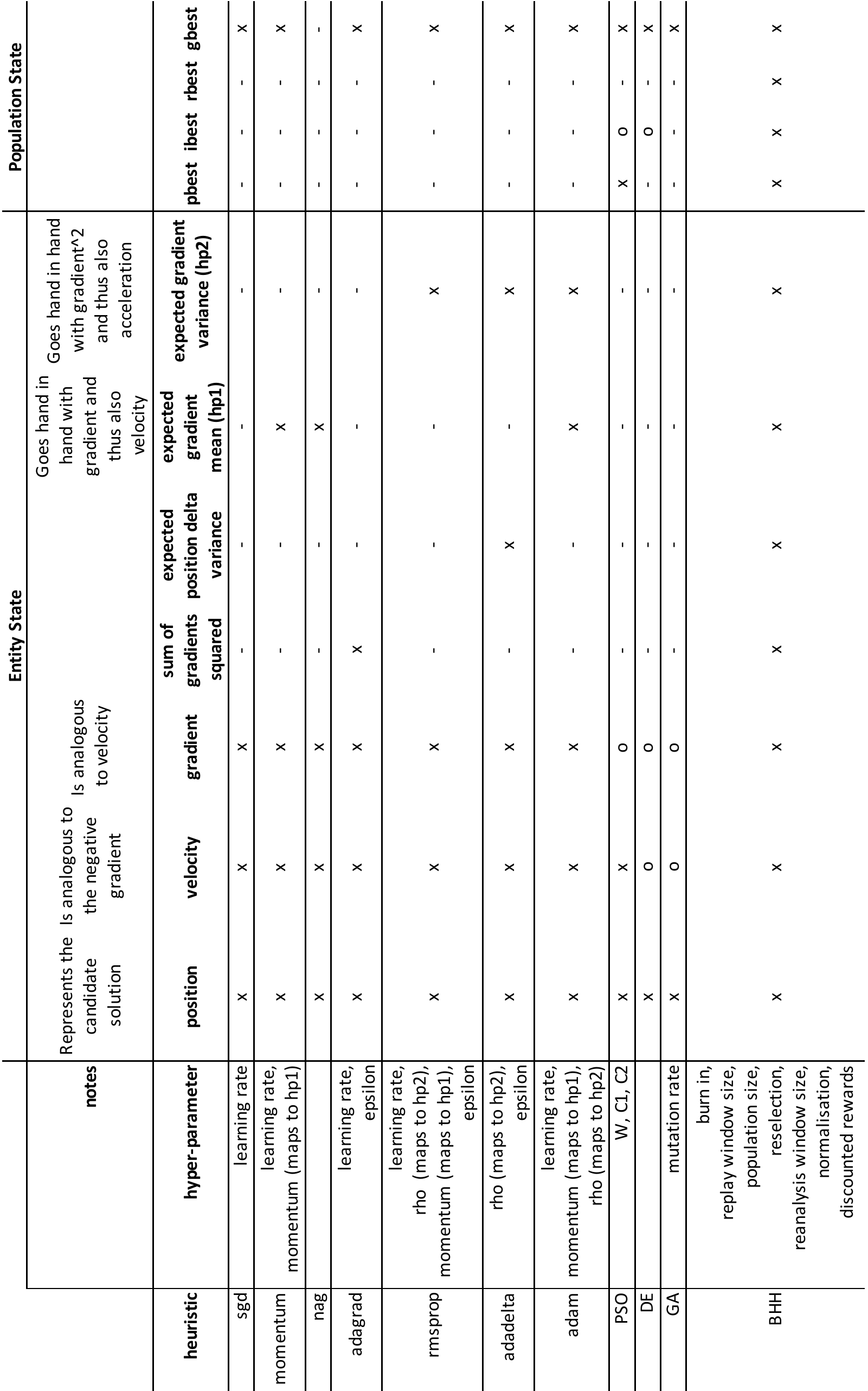}
	\caption{Mapping of proxied heuristic state update operations as implemented by the \acs{BHH}}
	\label{fig:methodology:heuristics:proxies}%
\end{figure}

The mapping of proxied heuristic state update operations implemented by the \acs{BHH} in the empirical process is given in Figure \ref{fig:methodology:heuristics:proxies}. In Figure \ref{fig:methodology:heuristics:proxies}, cells containing \textbf{x} indicate that the associated heuristic implements that particular state parameter explicitly, and cells containing \textbf{o} indicate that the state parameter is implicitly implemented as part of the \acs{BHH} algorithm.

\subsection{Datasets}\label{sec:methodology:datasets}

In the context of training \acp{FFNN}, the underlying models are trained across a number of datasets. All the datasets used in the empirical process originate from the UCI Machine Learning Repository \citep{ref:uci:2022}. Datasets are grouped by problem type and include seven classification and seven regression datasets. The details around the datasets used can be found in Tables \ref{tab:methodology:datasets:classification} and \ref{tab:methodology:datasets:regression}. Each dataset is split into a training set comprising 80\% of the data, and a test set comprising 20\% of the data.

A number of classification datasets contain an unbalanced representation of classes. This work does not apply mechanisms to cater for class balancing, in order to eliminate as many variables and factors in the empirical process as possible.

\begin{table}[!tb]
	\centering
	\caption{Classification datasets}
	\label{tab:methodology:datasets:classification}%
	\par\bigskip
	\resizebox{\textwidth}{!}{
		\begin{tabular}{ccccccccc}
			\textbf{dataset} & \textbf{output} & \textbf{types}             & \textbf{attributes} & \textbf{classes} & \textbf{instances} & \textbf{batch} & \textbf{steps} & \textbf{citation}          \\
			\midrule
			iris             & multivariate    & real                       & 4                   & 3                & 150                & 16             & 10             & \citep{ref:fisher:1936}    \\
			car              & multivariate    & categorical                & 6                   & 4                & 1728               & 128            & 14             & \citep{ref:bohanec:1988}   \\
			abalone          & multivariate    & categorical, integer, real & 8                   & 28               & 4177               & 256            & 17             & \citep{ref:waugh:1995}     \\
			wine quality     & multivariate    & real                       & 12                  & 11               & 4898               & 256            & 20             & \citep{ref:cortez:2009}    \\
			mushroom         & multivariate    & categorical                & 22                  & 2                & 8214               & 512            & 17             & \citep{ref:schlimmer:1987} \\
			bank             & multivariate    & real                       & 17                  & 2                & 45211              & 512            & 89             & \citep{ref:moro:2014}      \\
			diabetic         & multivariate    & integer                    & 55                  & 3                & 100000             & 1024           & 98             & \citep{ref:strack:2014}    \\
		\end{tabular}%
	}
\end{table}%

\begin{table}[!tb]
	\centering
	\caption{Regression datasets}
	\label{tab:methodology:datasets:regression}%
	\par\bigskip
	\resizebox{\textwidth}{!}{
		\begin{tabular}{cccccccc}
			\textbf{dataset}    & \textbf{output}           & \textbf{types} & \textbf{attributes} & \textbf{instances} & \textbf{batch} & \textbf{steps} & \textbf{citation}         \\
			\midrule
			fish toxicity       & multivariate              & real           & 7                   & 908                & 64             & 15             & \citep{ref:cassotti:2015} \\
			housing             & univariate                & real           & 13                  & 506                & 32             & 16             & \citep{ref:harrison:1978} \\
			forest fires        & multivariate              & real           & 13                  & 517                & 32             & 17             & \citep{ref:cortez:2007}   \\
			student performance & multivariate              & integer        & 33                  & 649                & 32             & 21             & \citep{ref:cortez:2008}   \\
			parkinsons          & multivariate              & integer, real  & 26                  & 5875               & 256            & 23             & \citep{ref:tsanas:2009}   \\
			air quality         & multivariate, time series & real           & 15                  & 9358               & 256            & 37             & \citep{ref:de:2008}       \\
			bike                & univariate                & integer, real  & 16                  & 17389              & 256            & 68             & \citep{ref:fanaee:2014}   \\
		\end{tabular}%
	}
\end{table}%

\subsection{Models}\label{sec:methodology:model}

All models trained in the empirical process follow implementations of shallow \acp{FFNN} with only one hidden layer. The number of hidden units used were determined empirically. Weights are initialised by means of \textit{Glorot uniform sampling} \citep{ref:glorot:2010}. The models and their configuration, as it is used for each dataset, are given in Table \ref{tab:methodology:models:configurations}.

\begin{table}[!tb]
	\centering
	\caption{Model configurations}
	\label{tab:methodology:models:configurations}%
	\par\bigskip
	\resizebox{\textwidth}{!}{
		\begin{tabular}{rcccccccc}
			\textbf{dataset}    & \textbf{inputs} & \textbf{hidden} & \textbf{output} & \textbf{biases} & \textbf{parameters} & \textbf{topology} & \textbf{l1 activation} & \textbf{l2 activation} \\
			\midrule
			fish toxicity       & 6               & 3               & 1               & yes             & 25                  & dense             & LReLU ($\alpha = 0.3$) & sigmoid                \\
			iris                & 4               & 5               & 3               & yes             & 43                  & dense             & LReLU ($\alpha = 0.3$) & softmax                \\
			air quality         & 12              & 8               & 1               & yes             & 113                 & dense             & LReLU ($\alpha = 0.3$) & sigmoid                \\
			housing             & 13              & 8               & 1               & yes             & 121                 & dense             & LReLU ($\alpha = 0.3$) & sigmoid                \\
			wine quality        & 13              & 10              & 7               & yes             & 217                 & dense             & LReLU ($\alpha = 0.3$) & softmax                \\
			parkinsons          & 21              & 10              & 1               & yes             & 231                 & dense             & LReLU ($\alpha = 0.3$) & sigmoid                \\
			car                 & 21              & 10              & 4               & yes             & 264                 & dense             & LReLU ($\alpha = 0.3$) & softmax                \\
			forest fires        & 43              & 16              & 1               & yes             & 721                 & dense             & LReLU ($\alpha = 0.3$) & sigmoid                \\
			abalone             & 10              & 36              & 28              & yes             & 1432                & dense             & LReLU ($\alpha = 0.3$) & softmax                \\
			bank                & 51              & 32              & 1               & yes             & 1697                & dense             & LReLU ($\alpha = 0.3$) & softmax                \\
			bike                & 61              & 32              & 1               & yes             & 2017                & dense             & LReLU ($\alpha = 0.3$) & sigmoid                \\
			student performance & 99              & 32              & 1               & yes             & 3233                & dense             & LReLU ($\alpha = 0.3$) & sigmoid                \\
			adult               & 108             & 64              & 1               & yes             & 7041                & dense             & LReLU ($\alpha = 0.3$) & softmax                \\
			mushroom            & 117             & 64              & 1               & yes             & 7617                & dense             & LReLU ($\alpha = 0.3$) & softmax                \\
			diabetic            & 2369            & 32              & 3               & yes             & 75939               & dense             & LReLU ($\alpha = 0.3$) & softmax                \\
		\end{tabular}%
	}
\end{table}%

\subsection{Performance Measures}\label{sec:methodology:performance_measures}

\Acf{BinXE} is used for classification problems with two classes and \acf{SparseCatXE} is used for classification problems with more than two classes. For the classification problems, accuracy is also measured. For regression problems, the \acf{MSE} is used as a performance metric. After training has completed, the \textit{average rank}, based on test loss, for all configurations, is calculated at each mini-batch step.

\subsection{Statistical Analysis}
\label{sec:methodology:statistical_analysis}

Each experiment and configuration is trained for a maximum of 30 epochs and is repeated over 30 independent runs, for each of the datasets. No early-stopping mechanism is used. Statistical analysis is executed on the results from the test datasets. An average rank is calculated across all 30 runs, for both experimental groups and configurations, at each step, for every epoch of training.

The Shapiro-Wilk test for normality \citep{ref:shapiro:1965} ($\alpha$ = 0.001) is used to determine if the results are normally distributed. The Levene's test for equality of variance \citep{ref:levene:1961} ($\alpha$ = 0.001) is used. For experiments with three or more configurations, the \acs{ANOVA} statistical test \citep{ref:fisher:1921} ($\alpha$ = 0.001) is used. The Kruskal-Wallis ranked non-parametric test \citep{ref:kruskal:1952} for statistical significance ($\alpha$ = 0.001) is used for cases where data is not normally distributed. Finally, a post-hoc Tukey honest significant difference test \citep{ref:tukey:1949} ($\alpha$ = 0.001) is used from which significant ranking is retrieved. Descriptive and critical difference plots are then retrieved from these results to provide visual aid.

\subsection{Implementation}\label{sec:methodology:implementation}

All implementations are done from first principles in Python 3.9 using Tensorflow 2.7 and Tensorflow Probability 0.15.0. The source code and data for this research is provided and made public at \url{https://github.com/arneschreuder/masters}.

\section{Results}
\label{sec:results}

This section provides the results of the empirical process that has been conducted. Detailed discussions follow on the outcomes of each experiment.

\subsection{Overview}\label{sec:results:overview}

This section provides a brief discussion on the general outcome of the empirical process as a whole and identifies some key aspects to be kept in mind when interpreting the results of the experiments.

Firstly, the \acs{BHH} applies a form of online learning. As such, the \acs{BHH} applies the learning mechanism during training in a single run of the training process. The training process is not repeated iteratively as is the case with some \acp{HH}.

Most of the training progress is observed to occur within the first five epochs. As a result, the \acs{BHH} should apply most learning at the early stages of the training process. After five epochs, the training of most of the underlying \acp{FFNN} converges and little performance gain is observed after that point. Since this empirical process does not apply early stopping of the training process, the \acs{BHH} will continue to explore the heuristic space beyond the five epoch mark.

The \acs{BHH} does not implement a type of move-acceptance strategy where the application of a heuristic to an entity is only accepted if it leads to a better solution. In some cases, the \acs{BHH} then selects heuristics that yield sub-optimal results, but is shown to mostly return to optimal solutions over a number of steps.

Given the stochastic nature of the heuristic selection mechanism, sufficient samples of the performance of each heuristics-entity combination in the performance log are required for the \acs{BHH} to learn. This requirement is further strengthened by the Bayesian nature of the probabilistic model implemented by the \acs{BHH}. The probabilistic model implements \textit{probability distributions of heuristic selection probabilities} and as such, insufficient samples in the performance log could render a form of random search.

By default, the \acs{BHH} baseline configuration has a reanalysis interval of 10, and a replay window size of 10, which is a small window to learn from. Despite the small reanalysis interval and the small replay window size, it should be observed that the \acs{BHH} exploits small performance biases and finds small performance gains throughout.

\subsection{BHH vs. Low-Level Heuristics}\label{sec:results:standalone}

Tables \ref{tab:results:standalone:metrics:rank_a} and \ref{tab:results:standalone:metrics:rank_b} provide the empirical results in ranked format. The performance rank is calculated as the average rank produced by each heuristic, across all datasets, for all independent runs and all epochs. The average rank across all epochs produces a view on the performance of the heuristics as it relates to the entire training process. Finally, a normalised average rank is provided for the overall performance of all heuristics at the bottom of the table. The normalised average rank is calculated as a discrete normalisation of the average rank achieved across all datasets, for all independent runs and epochs.

Tables \ref{tab:results:standalone:metrics:rank_a} and \ref{tab:results:standalone:metrics:rank_b} show that the \textit{bhh\_gd} configuration produced the best results of the \acs{BHH} variants and managed to perform well, producing generally good results across all datasets. The \textit{bhh\_gd} configuration managed to produce results that are comparable to the top three heuristics for each dataset, while the \textit{bhh\_all} and \textit{bhh\_mh} produced average results compared to all the heuristics.

\begin{table}[!tb]
	\centering
	\caption{Empirical results showing normalised average rank and statistics for the top six low-level heuristics and three heuristic pool variants of the \acs{BHH} baseline configuration, across multiple datasets, for all independent runs and epochs.}
	\label{tab:results:standalone:metrics:rank_a}%
	\par\bigskip
	\resizebox{\textwidth}{!}{
		\begin{tabular}{r|ccc|c|c|c|}
			                              & \multicolumn{6}{c|}{\textbf{BHH vs. Low-Level Heuristics - Average Rank (Part A)}}                                                                                                                                                                                                                                                                                                   \\
			\midrule
			\textbf{dataset}              & \textbf{adagrad}                                                                   & \textbf{adam}                                           & \textbf{rmsprop}                                        & \textbf{bhh\_gd}                                        & \textbf{nag}                                            & \textbf{bhh\_all}                                       \\
			\midrule
			\textbf{abalone}              & \cellcolor[rgb]{ .388,  .745,  .482}2,2215 ($\pm$1,591)                            & \cellcolor[rgb]{ .416,  .753,  .482}2,3989 ($\pm$1,887) & \cellcolor[rgb]{ .78,  .859,  .502}4,6172 ($\pm$2,65)   & \cellcolor[rgb]{ .796,  .863,  .506}4,7032 ($\pm$2,108) & \cellcolor[rgb]{ .725,  .839,  .498}4,2731 ($\pm$1,542) & \cellcolor[rgb]{ 1,  .922,  .518}5,9376 ($\pm$2,399)    \\
			\textbf{air\_quality}         & \cellcolor[rgb]{ .427,  .757,  .482}3,6409 ($\pm$2,259)                            & \cellcolor[rgb]{ .804,  .863,  .506}5,4312 ($\pm$2,62)  & \cellcolor[rgb]{ .388,  .745,  .482}3,4452 ($\pm$2,57)  & \cellcolor[rgb]{ .729,  .843,  .502}5,0817 ($\pm$2,762) & \cellcolor[rgb]{ .467,  .765,  .486}3,8194 ($\pm$2,229) & \cellcolor[rgb]{ 1,  .894,  .514}6,686 ($\pm$3,061)     \\
			\textbf{bank}                 & \cellcolor[rgb]{ .455,  .765,  .486}2,5495 ($\pm$1,598)                            & \cellcolor[rgb]{ .388,  .745,  .482}2,0796 ($\pm$1,587) & \cellcolor[rgb]{ .588,  .8,  .49}3,4645 ($\pm$2,209)    & \cellcolor[rgb]{ .8,  .863,  .506}4,8828 ($\pm$1,702)   & \cellcolor[rgb]{ .71,  .835,  .498}4,2871 ($\pm$1,732)  & \cellcolor[rgb]{ 1,  .922,  .518}6,2419 ($\pm$2,157)    \\
			\textbf{bike}                 & \cellcolor[rgb]{ .388,  .745,  .482}1,7204 ($\pm$1,384)                            & \cellcolor[rgb]{ .643,  .816,  .494}3,6925 ($\pm$4,004) & \cellcolor[rgb]{ .973,  .914,  .514}6,2624 ($\pm$4,58)  & \cellcolor[rgb]{ .663,  .824,  .498}3,8441 ($\pm$1,398) & \cellcolor[rgb]{ 1,  .922,  .518}6,4516 ($\pm$1,02)     & \cellcolor[rgb]{ .71,  .835,  .498}4,2151 ($\pm$1,361)  \\
			\textbf{car}                  & \cellcolor[rgb]{ .702,  .835,  .498}4,7634 ($\pm$0,938)                            & \cellcolor[rgb]{ .388,  .745,  .482}1,6226 ($\pm$1,405) & \cellcolor[rgb]{ .455,  .765,  .486}2,3269 ($\pm$1,409) & \cellcolor[rgb]{ .561,  .792,  .49}3,3473 ($\pm$1,35)   & \cellcolor[rgb]{ .831,  .871,  .506}6,0785 ($\pm$0,799) & \cellcolor[rgb]{ .58,  .8,  .49}3,5624 ($\pm$1,315)     \\
			\textbf{diabetic}             & \cellcolor[rgb]{ .506,  .776,  .486}2,7796 ($\pm$1,659)                            & \cellcolor[rgb]{ 1,  .886,  .514}7,1484 ($\pm$2,227)    & \cellcolor[rgb]{ 1,  .922,  .518}6,7376 ($\pm$2,577)    & \cellcolor[rgb]{ .812,  .867,  .506}5,2269 ($\pm$2,186) & \cellcolor[rgb]{ .388,  .745,  .482}1,8118 ($\pm$1,413) & \cellcolor[rgb]{ .988,  .69,  .475}9,3968 ($\pm$3,022)  \\
			\textbf{fish\_toxicity}       & \cellcolor[rgb]{ .533,  .784,  .49}4,2645 ($\pm$2,614)                             & \cellcolor[rgb]{ .388,  .745,  .482}3,6022 ($\pm$2,445) & \cellcolor[rgb]{ .388,  .745,  .482}3,5946 ($\pm$2,329) & \cellcolor[rgb]{ .784,  .859,  .502}5,4118 ($\pm$2,665) & \cellcolor[rgb]{ .89,  .89,  .51}5,8914 ($\pm$2,629)    & \cellcolor[rgb]{ .875,  .886,  .51}5,829 ($\pm$2,856)   \\
			\textbf{forest\_fires}        & \cellcolor[rgb]{ .769,  .855,  .502}5,1559 ($\pm$2,922)                            & \cellcolor[rgb]{ .388,  .745,  .482}4,2688 ($\pm$2,984) & \cellcolor[rgb]{ .718,  .839,  .498}5,0355 ($\pm$3,143) & \cellcolor[rgb]{ .569,  .796,  .49}4,6935 ($\pm$2,759)  & \cellcolor[rgb]{ 1,  .922,  .518}5,6882 ($\pm$2,215)    & \cellcolor[rgb]{ .91,  .894,  .51}5,4839 ($\pm$3,107)   \\
			\textbf{housing}              & \cellcolor[rgb]{ .404,  .749,  .482}3,4484 ($\pm$2,025)                            & \cellcolor[rgb]{ .388,  .745,  .482}3,3344 ($\pm$1,819) & \cellcolor[rgb]{ .439,  .757,  .482}3,6946 ($\pm$2,166) & \cellcolor[rgb]{ .553,  .792,  .49}4,4742 ($\pm$2,312)  & \cellcolor[rgb]{ .584,  .8,  .49}4,6839 ($\pm$2,658)    & \cellcolor[rgb]{ .537,  .788,  .49}4,3763 ($\pm$2,438)  \\
			\textbf{iris}                 & \cellcolor[rgb]{ .965,  .91,  .514}6,3946 ($\pm$1,6)                               & \cellcolor[rgb]{ .525,  .784,  .49}3,5839 ($\pm$2,511)  & \cellcolor[rgb]{ .388,  .745,  .482}2,6968 ($\pm$1,912) & \cellcolor[rgb]{ .706,  .835,  .498}4,7473 ($\pm$2,275) & \cellcolor[rgb]{ .522,  .78,  .486}3,5548 ($\pm$2,125)  & \cellcolor[rgb]{ .78,  .859,  .502}5,2204 ($\pm$3,041)  \\
			\textbf{mushroom}             & \cellcolor[rgb]{ .71,  .835,  .498}4,4656 ($\pm$1,053)                             & \cellcolor[rgb]{ .388,  .745,  .482}2,1344 ($\pm$1,883) & \cellcolor[rgb]{ .431,  .757,  .482}2,4656 ($\pm$1,359) & \cellcolor[rgb]{ .569,  .796,  .49}3,4484 ($\pm$1,602)  & \cellcolor[rgb]{ .969,  .91,  .514}6,3323 ($\pm$0,891)  & \cellcolor[rgb]{ .6,  .804,  .494}3,6688 ($\pm$2,469)   \\
			\textbf{parkinsons}           & \cellcolor[rgb]{ .412,  .749,  .482}2,4677 ($\pm$1,497)                            & \cellcolor[rgb]{ .388,  .745,  .482}2,2333 ($\pm$1,742) & \cellcolor[rgb]{ .541,  .788,  .49}3,5656 ($\pm$2,492)  & \cellcolor[rgb]{ .655,  .82,  .494}4,572 ($\pm$1,934)   & \cellcolor[rgb]{ 1,  .922,  .518}7,5355 ($\pm$1,44)     & \cellcolor[rgb]{ .635,  .816,  .494}4,3839 ($\pm$1,861) \\
			\textbf{student\_performance} & \cellcolor[rgb]{ .388,  .745,  .482}2,5634 ($\pm$1,912)                            & \cellcolor[rgb]{ .98,  .506,  .439}11,3978 ($\pm$2,178) & \cellcolor[rgb]{ .973,  .412,  .42}12,4312 ($\pm$1,34)  & \cellcolor[rgb]{ .843,  .875,  .506}5,6624 ($\pm$3,57)  & \cellcolor[rgb]{ .478,  .769,  .486}3,1935 ($\pm$2,12)  & \cellcolor[rgb]{ .875,  .882,  .51}5,8634 ($\pm$3,159)  \\
			\textbf{wine\_quality}        & \cellcolor[rgb]{ .569,  .796,  .49}3,2806 ($\pm$1,931)                             & \cellcolor[rgb]{ .388,  .745,  .482}2,1118 ($\pm$1,666) & \cellcolor[rgb]{ .624,  .812,  .494}3,6301 ($\pm$1,731) & \cellcolor[rgb]{ .808,  .863,  .506}4,7882 ($\pm$2,105) & \cellcolor[rgb]{ .706,  .835,  .498}4,1505 ($\pm$1,916) & \cellcolor[rgb]{ .871,  .882,  .51}5,1925 ($\pm$1,951)  \\
			\midrule
			\textbf{avg rank}             & \cellcolor[rgb]{ .388,  .745,  .482}3,5512 ($\pm$2,25)                             & \cellcolor[rgb]{ .471,  .769,  .486}3,9314 ($\pm$3,423) & \cellcolor[rgb]{ .616,  .808,  .494}4,5691 ($\pm$3,517) & \cellcolor[rgb]{ .631,  .812,  .494}4,6346 ($\pm$2,364) & \cellcolor[rgb]{ .675,  .827,  .498}4,8394 ($\pm$2,384) & \cellcolor[rgb]{ .808,  .867,  .506}5,4327 ($\pm$2,9)   \\
			\midrule
			\textbf{normalised avg rank}  & \cellcolor[rgb]{ .388,  .745,  .482}1                                              & \cellcolor[rgb]{ .49,  .773,  .486}2                    & \cellcolor[rgb]{ .592,  .804,  .494}3                   & \cellcolor[rgb]{ .694,  .831,  .498}4                   & \cellcolor[rgb]{ .796,  .863,  .506}5                   & \cellcolor[rgb]{ .898,  .89,  .51}6                     \\
			\cmidrule{5-5}\cmidrule{7-7}\end{tabular}%
	}
\end{table}%

\begin{table}[!tb]
	\centering
	\caption{Empirical results showing normalised average rank and statistics for the bottom six low-level heuristics and three heuristic pool variants of the \acs{BHH} baseline configuration, across multiple datasets, for all independent runs and epochs.}
	\label{tab:results:standalone:metrics:rank_b}%
	\par\bigskip
	\resizebox{\textwidth}{!}{
		\begin{tabular}{r|c|c|ccccc|}
			                              & \multicolumn{7}{c}{\textbf{BHH vs. Low-Level Heuristics - Average Rank (Part B)}}                                                                                                                                                                                                                                                                                                                                                                  \\
			\midrule
			\textbf{dataset}              & \textbf{adadelta}                                                                 & \textbf{bhh\_mh}                                        & \textbf{ga}                                              & \textbf{pso}                                             & \textbf{sgd}                                             & \textbf{momentum}                                        & \textbf{de}                                              \\
			\midrule
			\textbf{abalone}              & \cellcolor[rgb]{ .894,  .89,  .51}5,3129 ($\pm$1,478)                             & \cellcolor[rgb]{ .992,  .749,  .486}8,1882 ($\pm$1,195) & \cellcolor[rgb]{ .98,  .525,  .443}11,1108 ($\pm$1,102)  & \cellcolor[rgb]{ .98,  .514,  .439}11,2559 ($\pm$1,826)  & \cellcolor[rgb]{ .992,  .718,  .478}8,628 ($\pm$1,019)   & \cellcolor[rgb]{ .984,  .624,  .463}9,8151 ($\pm$1,16)   & \cellcolor[rgb]{ .973,  .412,  .42}12,5376 ($\pm$1,329)  \\
			\textbf{air\_quality}         & \cellcolor[rgb]{ .765,  .851,  .502}5,2441 ($\pm$3,162)                           & \cellcolor[rgb]{ 1,  .922,  .518}6,357 ($\pm$2,303)     & \cellcolor[rgb]{ .996,  .784,  .494}7,8204 ($\pm$2,265)  & \cellcolor[rgb]{ .984,  .588,  .455}9,9151 ($\pm$2,288)  & \cellcolor[rgb]{ .98,  .518,  .443}10,6613 ($\pm$1,606)  & \cellcolor[rgb]{ .973,  .412,  .42}11,7559 ($\pm$1,473)  & \cellcolor[rgb]{ .976,  .471,  .431}11,1419 ($\pm$2,236) \\
			\textbf{bank}                 & \cellcolor[rgb]{ .914,  .894,  .51}5,672 ($\pm$1,241)                             & \cellcolor[rgb]{ .988,  .639,  .463}9,7495 ($\pm$1,048) & \cellcolor[rgb]{ .98,  .537,  .447}10,9817 ($\pm$1,216)  & \cellcolor[rgb]{ .976,  .467,  .431}11,8376 ($\pm$1,464) & \cellcolor[rgb]{ .992,  .757,  .486}8,2774 ($\pm$1,03)   & \cellcolor[rgb]{ .992,  .741,  .486}8,4774 ($\pm$1,068)  & \cellcolor[rgb]{ .973,  .412,  .42}12,4989 ($\pm$1,224)  \\
			\textbf{bike}                 & \cellcolor[rgb]{ .859,  .878,  .506}5,3602 ($\pm$1,155)                           & \cellcolor[rgb]{ .996,  .843,  .506}7,4108 ($\pm$1,008) & \cellcolor[rgb]{ .988,  .686,  .475}9,2269 ($\pm$1,183)  & \cellcolor[rgb]{ .988,  .682,  .475}9,3086 ($\pm$1,761)  & \cellcolor[rgb]{ .984,  .596,  .455}10,3355 ($\pm$1,419) & \cellcolor[rgb]{ .984,  .561,  .451}10,7086 ($\pm$1,423) & \cellcolor[rgb]{ .973,  .412,  .42}12,4634 ($\pm$1,465)  \\
			\textbf{car}                  & \cellcolor[rgb]{ 1,  .922,  .518}7,7344 ($\pm$1,746)                              & \cellcolor[rgb]{ .996,  .8,  .498}8,8505 ($\pm$1,413)   & \cellcolor[rgb]{ .984,  .592,  .455}10,7763 ($\pm$1,471) & \cellcolor[rgb]{ 1,  .855,  .506}8,3613 ($\pm$1,622)     & \cellcolor[rgb]{ .988,  .651,  .467}10,2452 ($\pm$1,447) & \cellcolor[rgb]{ .984,  .576,  .451}10,9226 ($\pm$1,349) & \cellcolor[rgb]{ .973,  .412,  .42}12,4086 ($\pm$1,492)  \\
			\textbf{diabetic}             & \cellcolor[rgb]{ .494,  .773,  .486}2,6753 ($\pm$1,629)                           & \cellcolor[rgb]{ .992,  .765,  .49}8,557 ($\pm$1,17)    & \cellcolor[rgb]{ .98,  .529,  .443}11,2011 ($\pm$1,413)  & \cellcolor[rgb]{ .984,  .573,  .451}10,7022 ($\pm$1,067) & \cellcolor[rgb]{ .898,  .89,  .51}5,9215 ($\pm$1,542)    & \cellcolor[rgb]{ .949,  .906,  .514}6,3355 ($\pm$1,612)  & \cellcolor[rgb]{ .973,  .412,  .42}12,5065 ($\pm$1,242)  \\
			\textbf{fish\_toxicity}       & \cellcolor[rgb]{ .996,  .788,  .494}7,914 ($\pm$3,429)                            & \cellcolor[rgb]{ 1,  .922,  .518}6,3849 ($\pm$2,944)    & \cellcolor[rgb]{ 1,  .894,  .514}6,7043 ($\pm$2,82)      & \cellcolor[rgb]{ .996,  .82,  .498}7,5731 ($\pm$2,982)   & \cellcolor[rgb]{ .976,  .471,  .431}11,5785 ($\pm$1,459) & \cellcolor[rgb]{ .973,  .412,  .42}12,2301 ($\pm$1,382)  & \cellcolor[rgb]{ .984,  .608,  .459}10,0215 ($\pm$2,358) \\
			\textbf{forest\_fires}        & \cellcolor[rgb]{ 1,  .859,  .506}6,5161 ($\pm$3,082)                              & \cellcolor[rgb]{ .898,  .89,  .51}5,4591 ($\pm$2,668)   & \cellcolor[rgb]{ .996,  .796,  .494}7,3667 ($\pm$2,37)   & \cellcolor[rgb]{ 1,  .863,  .51}6,4796 ($\pm$3,354)      & \cellcolor[rgb]{ .98,  .529,  .443}10,8129 ($\pm$1,207)  & \cellcolor[rgb]{ .976,  .463,  .431}11,7065 ($\pm$1,325) & \cellcolor[rgb]{ .973,  .412,  .42}12,3333 ($\pm$1,923)  \\
			\textbf{housing}              & \cellcolor[rgb]{ 1,  .918,  .518}7,5903 ($\pm$2,748)                              & \cellcolor[rgb]{ 1,  .922,  .518}7,5441 ($\pm$1,736)    & \cellcolor[rgb]{ 1,  .878,  .51}7,8839 ($\pm$2,099)      & \cellcolor[rgb]{ .984,  .608,  .459}9,9409 ($\pm$2,317)  & \cellcolor[rgb]{ .973,  .412,  .42}11,4075 ($\pm$1,528)  & \cellcolor[rgb]{ .976,  .431,  .424}11,2731 ($\pm$1,506) & \cellcolor[rgb]{ .976,  .42,  .424}11,3484 ($\pm$2,096)  \\
			\textbf{iris}                 & \cellcolor[rgb]{ .973,  .412,  .42}11,3527 ($\pm$1,779)                           & \cellcolor[rgb]{ 1,  .922,  .518}6,6075 ($\pm$2,555)    & \cellcolor[rgb]{ .992,  .749,  .486}8,2473 ($\pm$1,765)  & \cellcolor[rgb]{ .992,  .745,  .486}8,2731 ($\pm$4,384)  & \cellcolor[rgb]{ .98,  .518,  .443}10,3796 ($\pm$1,294)  & \cellcolor[rgb]{ .976,  .447,  .427}11,0548 ($\pm$1,409) & \cellcolor[rgb]{ .988,  .678,  .471}8,8871 ($\pm$3,251)  \\
			\textbf{mushroom}             & \cellcolor[rgb]{ 1,  .922,  .518}6,5538 ($\pm$1,071)                              & \cellcolor[rgb]{ .992,  .718,  .478}9,0452 ($\pm$1,093) & \cellcolor[rgb]{ .98,  .51,  .439}11,5731 ($\pm$1,193)   & \cellcolor[rgb]{ .996,  .816,  .498}7,872 ($\pm$0,92)    & \cellcolor[rgb]{ .988,  .659,  .471}9,7527 ($\pm$1,083)  & \cellcolor[rgb]{ .98,  .557,  .451}10,9785 ($\pm$1,108)  & \cellcolor[rgb]{ .973,  .412,  .42}12,7097 ($\pm$1,478)  \\
			\textbf{parkinsons}           & \cellcolor[rgb]{ .875,  .882,  .51}6,472 ($\pm$2,423)                             & \cellcolor[rgb]{ .996,  .843,  .506}8,3161 ($\pm$1,644) & \cellcolor[rgb]{ 1,  .898,  .514}7,7968 ($\pm$1,719)     & \cellcolor[rgb]{ .996,  .827,  .502}8,4892 ($\pm$1,901)  & \cellcolor[rgb]{ .98,  .494,  .435}11,7516 ($\pm$1,155)  & \cellcolor[rgb]{ .973,  .412,  .42}12,5419 ($\pm$1,351)  & \cellcolor[rgb]{ .984,  .584,  .455}10,8742 ($\pm$1,317) \\
			\textbf{student\_performance} & \cellcolor[rgb]{ .514,  .78,  .486}3,4194 ($\pm$2,006)                            & \cellcolor[rgb]{ 1,  .902,  .514}6,9333 ($\pm$2,44)     & \cellcolor[rgb]{ 1,  .886,  .514}7,1032 ($\pm$1,989)     & \cellcolor[rgb]{ .98,  .537,  .443}11,0624 ($\pm$1,067)  & \cellcolor[rgb]{ .988,  .918,  .514}6,6366 ($\pm$2,023)  & \cellcolor[rgb]{ 1,  .922,  .518}6,7011 ($\pm$2,242)     & \cellcolor[rgb]{ .996,  .804,  .498}8,0323 ($\pm$1,935)  \\
			\textbf{wine\_quality}        & \cellcolor[rgb]{ 1,  .922,  .518}6,0011 ($\pm$2,404)                              & \cellcolor[rgb]{ .988,  .647,  .467}9,5935 ($\pm$1,494) & \cellcolor[rgb]{ .984,  .588,  .455}10,3387 ($\pm$1,62)  & \cellcolor[rgb]{ .98,  .525,  .443}11,1602 ($\pm$1,773)  & \cellcolor[rgb]{ .992,  .722,  .482}8,6344 ($\pm$1,18)   & \cellcolor[rgb]{ .988,  .651,  .467}9,5269 ($\pm$1,341)  & \cellcolor[rgb]{ .973,  .412,  .42}12,5903 ($\pm$1,352)  \\
			\midrule
			\textbf{avg rank}             & \cellcolor[rgb]{ 1,  .922,  .518}6,2727 ($\pm$3,004)                              & \cellcolor[rgb]{ .992,  .776,  .49}7,7855 ($\pm$2,271)  & \cellcolor[rgb]{ .988,  .639,  .467}9,1522 ($\pm$2,48)   & \cellcolor[rgb]{ .984,  .612,  .459}9,4451 ($\pm$2,75)   & \cellcolor[rgb]{ .984,  .592,  .455}9,6445 ($\pm$2,214)  & \cellcolor[rgb]{ .98,  .529,  .443}10,2877 ($\pm$2,346)  & \cellcolor[rgb]{ .973,  .412,  .42}11,4538 ($\pm$2,354)  \\
			\midrule
			\textbf{normalised avg rank}  & \cellcolor[rgb]{ 1,  .922,  .518}7                                                & \cellcolor[rgb]{ .996,  .839,  .502}8                   & \cellcolor[rgb]{ .992,  .753,  .486}9                    & \cellcolor[rgb]{ .988,  .667,  .471}10                   & \cellcolor[rgb]{ .984,  .584,  .455}11                   & \cellcolor[rgb]{ .98,  .498,  .439}12                    & \cellcolor[rgb]{ .973,  .412,  .42}13                    \\
			\cmidrule{3-3}\end{tabular}%
	}
\end{table}%

The normalised average ranks provided in Tables \ref{tab:results:standalone:metrics:rank_a} and \ref{tab:results:standalone:metrics:rank_b} show that the \textit{bhh\_gd} configuration ranked fourth, while the \textit{bhh\_al} and \textit{bhh\_mh} configurations ranked sixth and eighth amongst all thirteen heuristic implementations respectively. These results show that the \acs{BHH} generally performs well, but is not able to outperform the best heuristic for each dataset.

\begin{figure}[htb]
	\centering
	\includegraphics[width=\textwidth]{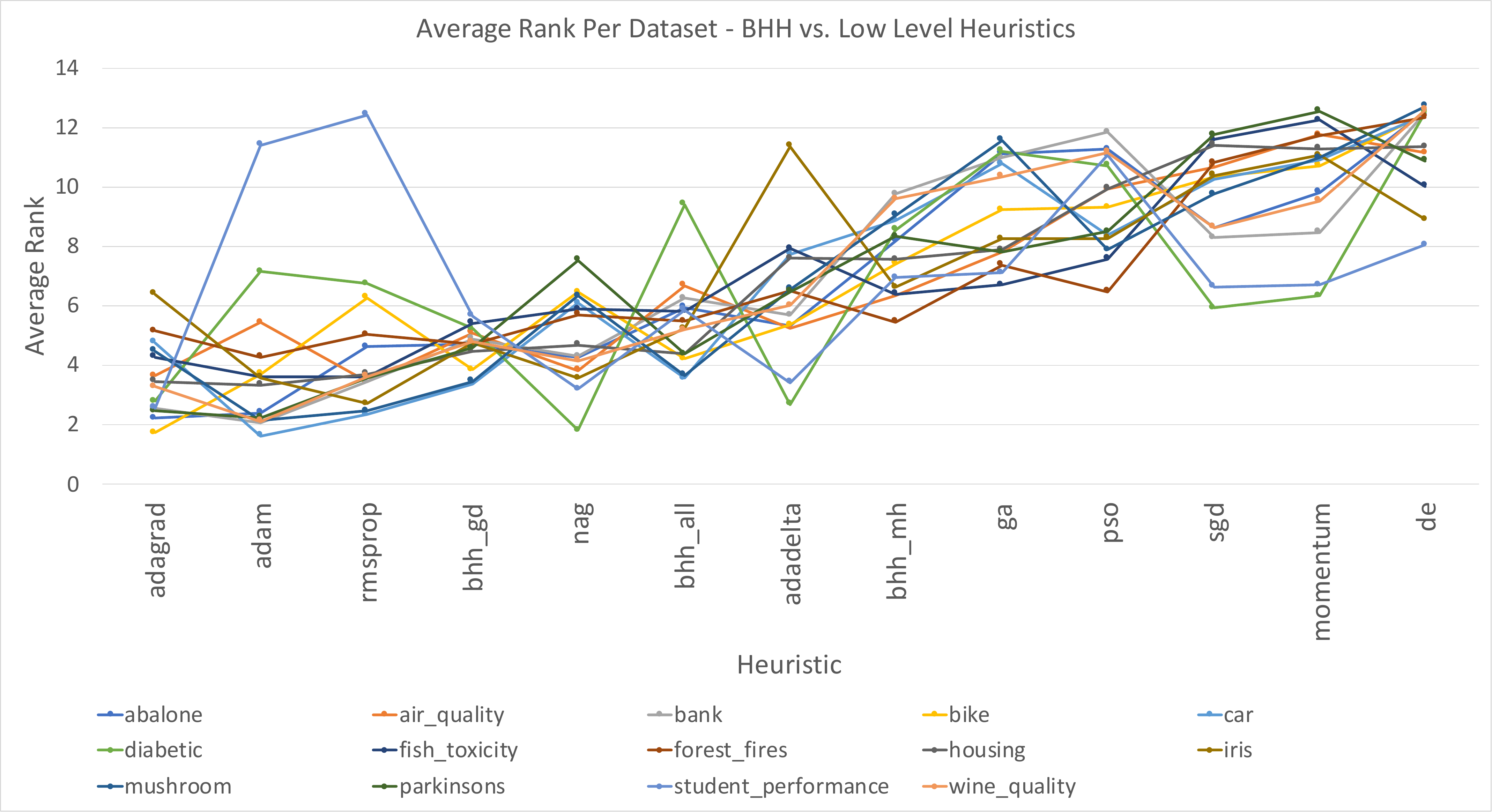}
	\caption{Descriptive plots for the average ranks of all low-level heuristics compared to three heuristic pool variants of the \acs{BHH} baseline configuration, per dataset, across all independent runs and epochs.}
	\label{fig:results:standalone:descriptive:descriptive}
\end{figure}

Figure \ref{fig:results:standalone:descriptive:descriptive} provides an illustration showing a descriptive plot of the average ranks achieved over all independent runs, for each heuristic, per dataset. The heuristics are ordered according to the normalised ranks presented in Tables \ref{tab:results:standalone:metrics:rank_a} and \ref{tab:results:standalone:metrics:rank_b}. Figure \ref{fig:results:standalone:descriptive:descriptive} shows that the \textit{bhh\_gd} heuristic achieved the lowest variance in average rank across all datasets, compared to the other heuristics. The aforementioned shows the generalisation capabilities of the \acs{BHH} to multiple problems.

\begin{figure}[htb]
	\centering
	\includegraphics[width=\textwidth]{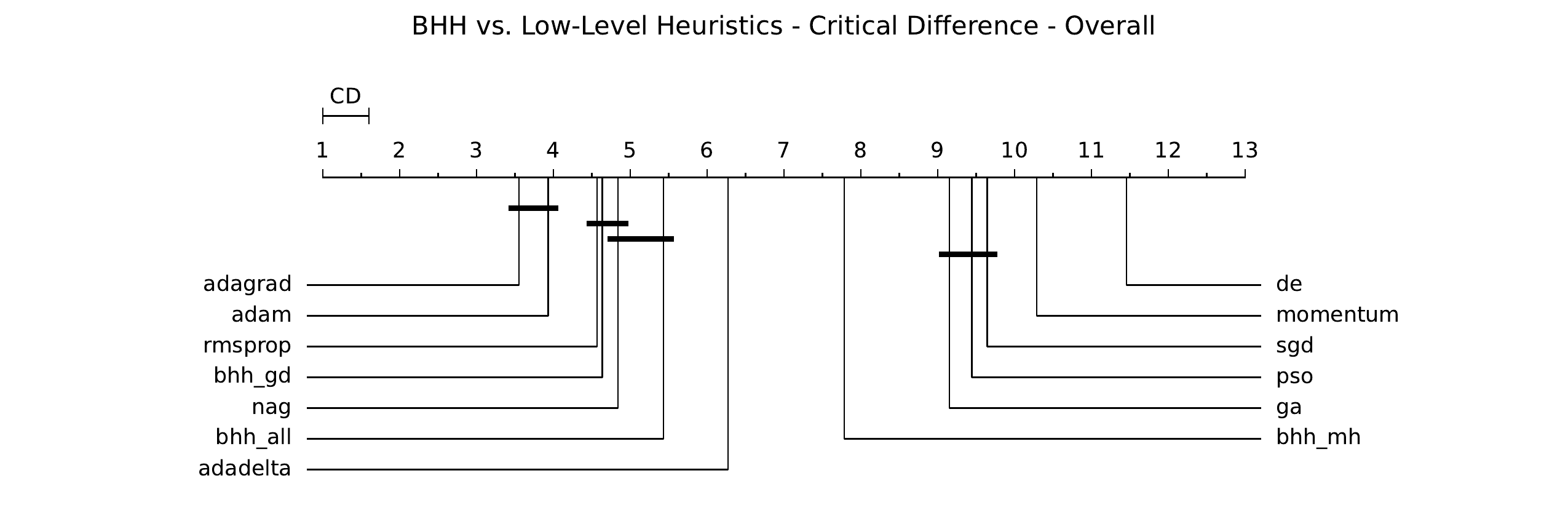}
	\caption{Critical difference plots for the average ranks of all low-level heuristics compared to three heuristic pool variants of the baseline \acs{BHH}, across all datasets, runs and epochs.}
	\label{fig:results:standalone:descriptive:cd}
\end{figure}

Figure \ref{fig:results:standalone:descriptive:cd} provides an illustration of the overall critical difference plots that illustrate the statistically significant differences in ranked performance for each heuristic as it relates to all datasets, across all independent runs and epochs. Although the outcomes of the \textit{bhh\_al} and \textit{bhh\_mh} configurations seem to produce average performance results, it should be noted that the performance difference between all heuristics is very small. Furthermore, the best configuration of the \acs{BHH}, namely the \textit{bhh\_gd} configuration, is statistically outperformed overall by only \acs{Adagrad} and \acs{Adam}, yielding statistically comparable results to \acs{RMSProp} and \acs{NAG}. It should be noted that the standalone low-level heuristics already produce good results in general across all datasets. In this particular case, producing better performance outcomes can be hard to achieve. However, as mentioned previously, the \acs{BHH} provides a generalisation capability across all datasets that is advantageous to the \acs{BHH}.

Another observation that can be made is that the gradient-based heuristics generally performed much better than the \acp{MH} on all datasets. State of the art methods for training \acp{FFNN}, such as \acs{Adam}, utilise gradient-based approaches that have been proven to work well on many occasions \citep{ref:kingma:2014}. Exploration of the heuristic space leads the \acs{BHH} to consider other heuristics during the training process, which could possibly result in worse performances at times. A suggestion to improve on these results is to include a move-acceptance strategy where heuristic progressions are discarded if they fail to produce better results.

\section{Conclusion}
\label{sec:conclusion}

The research done in this study stems from the difficult and tedious process of selecting the best heuristic for training \acp{FFNN}. The research presented in this article identified the possibility of using a different approach, referred to as \acp{HH}, to automate the heuristic selection process.

This research set out to develop a novel high-level heuristic that utilises probability theory in an online learning setting to drive the automatic heuristic selection process.

For the experimental group that compares the \acs{BHH} baseline with a number of low-level heuristics, it was found that the \textit{bhh\_gd} configuration, which contains only gradient-based heuristics in the heuristic pool, performed the best out of the \acs{BHH} variants, achieving an overall rank of fourth amongst thirteen heuristics that were implemented and executed on fourteen datasets. The \textit{bhh\_gd} configuration produced performance results close to that of the best low-level heuristics and was statistically outperformed only by the top two low-level heuristics. The \textit{bhh\_all} configuration, which contains only gradient-based heuristics and \acp{MH} in the heuristic pool, achieved an overall rank of sixth, and the \textit{bhh\_mh} configuration, which contains only \acp{MH} in the heuristic pool, achieved an overall rank of eighth.

Although the \textit{bhh\_gd} configuration produced performance results comparable to the best low-level heuristics, the \textit{bhh\_all} and \textit{bhh\_mh} configurations produced average results. It was found that, in general, gradient-based heuristics produced the best results, as such, it is understandable that the \textit{bhh\_gd} yielded the best performance outcomes between the different \acs{BHH} variants that were implemented. Although the \acs{BHH} variants were not able to produce better results than the top low-level heuristics, the \acs{BHH} variants still effectively trained the underlying \acp{FFNN} and produced good training outcomes overall. It was shown that the \textit{bhh\_gd} configuration produced the lowest variance in rank between datasets out of all of the heuristics implemented, giving the \acs{BHH} the ability to generalise well to other problems.

Finally, it was shown that the \acs{BHH} provides a mechanism whereby prior expert knowledge can be injected, before training starts. Future research can exploit this knowledge and provide a significant bias towards heuristics that are known to perform well on particular problem types. Future research can also investigate the scalability and effectiveness of the \acs{BHH} on other model architectures such as \acfp{DNN}. Furthermore, the selection mechanism of the \acs{BHH} can be extended to not just select heuristics from a heuristic pool, but also different model architectures from a model architecture pool.

\appendix

\section{Naïve Bayes}
\label{app:naive_bayes}

This section aims to dissect the probabilistic model that is presented in Equation~\eqref{eq:bhh:selection_mechanism:predictive_model_prop_to}. The \acs{BHH} implements a form of Naïve Bayes classifier, and thus independence between events can be assumed. The following derived \acp{PMF} are provided as fundamental building blocks to show the mechanism by which the \acs{BHH} learns.

The independence between events for class label $\boldsymbol{H}$, simply yields the \ac{PMF} of the \index{Multinomial probability distribution}Multinomial distribution as presented below:

\begin{equation}
	\label{eq:bhh:selection_mechanism:naive_bayes:h_pmf}
	\begin{split}
		P(\boldsymbol{H} \vert \boldsymbol{\theta})
		&\propto \prod_{i=1}^{I} \prod_{k=1}^{K} P(h_{i,k} \vert \theta_{k}) \\
		&\propto \prod_{i=1}^{I} \prod_{k=1}^{K} \theta_{k}^{\mathbbm{1}_{1}(h_{i,k})} \\
		&\propto \prod_{k=1}^{K} \theta_{k}^{\sum_{i=1}^{I} \mathbbm{1}_{1}(h_{i,k})} \\
		&\propto \prod_{k=1}^{K} \theta_{k}^{N_{k}}
	\end{split}
\end{equation}
\noindent
where $N_{k}$ is a summary variable such that $N_{k} = \sum_{i=i}^{I} \mathbbm{1}_{1}(h_{i,k})$, denoting the count of occurrences of the event $h_{i}$ taking on class $k$ in $I$ independent, identical runs.

The independence between events $\boldsymbol{E}$, given class label $\boldsymbol{H}$, is denoted by the likelihood of $\boldsymbol{E}$, conditional to the occurrence of heuristic $k$ and model parameter $\boldsymbol{\phi}$ as follows:

\begin{equation}
	\label{eq:bhh:selection_mechanism:naive_bayes:EgH_pmf}
	\begin{split}
		P(\boldsymbol{E} \vert \boldsymbol{H};  \boldsymbol{\phi})
		&\propto \prod_{i=1}^{I} \prod_{j=1}^{J} \prod_{k=1}^{K} P(e_{i,j,k} \vert h_{i,k} ; \phi_{j,k})  \\
		&\propto \prod_{i=1}^{I} \prod_{j=1}^{J} \prod_{k=1}^{K} \phi_{j,k}^{\mathbbm{1}_{1}(e_{i,j,k})\mathbbm{1}_{1}(h_{i,k})} \\
		&\propto \prod_{j=1}^{J} \prod_{k=1}^{K} \phi_{j,k}^{ \sum_{i}^{I} \left[ \mathbbm{1}_{1}(e_{i,j,k}) \mathbbm{1}_{1}(h_{i,k}) \right]} \\
		&\propto \prod_{j=1}^{J} \prod_{k=1}^{K} \phi_{j,k}^{N_{j,k}}
	\end{split}
\end{equation}
\noindent
where $N_{j,k}$ is a summary variable such that $N_{j,k} = \sum_{i=i}^{I} \mathbbm{1}_{1}(e_{i,j,k})\mathbbm{1}_{1}(h_{i,k})$, denoting the count of occurrences of the events $e_{i}$ taking on class $j$ and $h_{i}$ taking on class $k$, i.e. the count of occurrences of both entity $j$ and heuristic $k$ occurring together in $I$ independent, identical runs.

Finally, the independence between events for the performance criteria $\boldsymbol{C}$, given class label $\boldsymbol{H}$, is denoted by the likelihood of $\boldsymbol{C}$, conditional to the occurrence of heuristic $k$ and model parameter $\boldsymbol{\psi}$ as given below:

\begin{equation}
	\label{eq:bhh:selection_mechanism:naive_bayes:CgH_pmf}
	\begin{split}
		P(\boldsymbol{C} \vert \boldsymbol{H}; \boldsymbol{\psi})
		&\propto\prod_{i=1}^{I} \prod_{k=1}^{K} P(c_{i,k} \vert h_{i,k} ; \psi_{k})  \\
		&\propto \prod_{i=1}^{I} \prod_{k=1}^{K} \psi_{k}^{\mathbbm{1}_{1}(c_{i,k})\mathbbm{1}_{1}(h_{i,k})} (1 - \psi_{k})^{\mathbbm{1}_{0}(c_{i,k})\mathbbm{1}_{1}(h_{i,k})}\\
		&\propto \prod_{k=1}^{K} \psi_{k}^{\sum_{i=1}^{I} \mathbbm{1}_{1}(c_{i,k})\mathbbm{1}_{1}(h_{i,k})} (1 - \psi_{k})^{\sum_{i=1}^{I} \mathbbm{1}_{0}(c_{i,k})\mathbbm{1}_{1}(h_{i,k})}\\
		&\propto \prod_{k=1}^{K} \psi_{k}^{N_{1,k}} (1 - \psi_{k})^{N_{0,k}} \\
		&\propto \prod_{k=1}^{K} \psi_{k}^{N_{1,k}} (1 - \psi_{k})^{(N_{k} - N_{1,k})}
	\end{split}
\end{equation}
\noindent
where $N_{k}$ is the same summary variable as described for Equation~\eqref{eq:bhh:selection_mechanism:naive_bayes:h_pmf}. $N_{1,k}$ is a summary variable such that $N_{1,k} = \sum_{i=1}^{I} \mathbbm{1}_{1}(c_{i,k})\mathbbm{1}_{1}(h_{i,k})$, denoting the count of occurrences of the events $c_{i}$ taking on a success (i.e. $c_{i}=1$) and $h_{i}$ taking on class $k$, i.e. the count of occurrences of both succeeding in the performance criteria and heuristic $k$ occurring together in $I$ independent, identical runs. Similarly, $N_{0,k} = N_{k} - N_{1,k}$ denotes the count of occurrences of the events $c_{i}$ taking on a failure (i.e. $c_{i}=0$) and $h_{i}$ taking on class $k$.

Equations~\eqref{eq:bhh:selection_mechanism:naive_bayes:h_pmf}, \eqref{eq:bhh:selection_mechanism:naive_bayes:EgH_pmf} and \eqref{eq:bhh:selection_mechanism:naive_bayes:CgH_pmf} can be substituted into the proportional evaluation of the predictive model as given in Equation~\eqref{eq:bhh:selection_mechanism:predictive_model_prop_to}, resulting in

\begin{equation}
	\label{eq:bhh:selection_mechanism:naive_bayes:HgEC_pmf}
	\begin{split}
		P(\boldsymbol{H} \vert \boldsymbol{E}, \boldsymbol{C};  \boldsymbol{\theta}, \boldsymbol{\phi}, \boldsymbol{\psi})
		&\propto P(\boldsymbol{E} \vert \boldsymbol{H};  \boldsymbol{\phi})  P(\boldsymbol{C} \vert \boldsymbol{H}; \boldsymbol{\psi}) P(\boldsymbol{H} \vert \boldsymbol{\theta})  \\
		&\propto \left[ \prod_{j=1}^{J} \prod_{k=1}^{K} \phi_{j,k}^{N_{j,k}} \right] \left[ \prod_{k=1}^{K} \psi_{k}^{N_{1,k}} (1 - \psi_{k})^{(N_{k} - N_{1,k})} \right] \left[ \prod_{k=1}^{K} \theta_{k}^{N_{k}} \right]
	\end{split}
\end{equation}

Consider the practical implementation of the predictive model as shown in Equation~\eqref{eq:bhh:selection_mechanism:naive_bayes:HgEC_pmf}. Computationally, the equation presented in Equation~\eqref{eq:bhh:selection_mechanism:naive_bayes:HgEC_pmf} will underflow on a real computer if the resulting probabilities are very small.

\subsection{Numerical Stability}\label{sec:bhh:selection_mechanism:numerical_stability}

When Equation~\eqref{eq:bhh:selection_mechanism:naive_bayes:HgEC_pmf} is evaluated, the numerical stability is shown to underflow if the resulting probabilities from its parts are very small. Multiplication of multiple fractional parameters leads to an even smaller fractional number. Probabilities might be very low at some points during training. A solution to the aforementioned problem is to apply the \textit{log-sum-exp} trick. The transformation of Equation~\eqref{eq:bhh:selection_mechanism:naive_bayes:HgEC_pmf} using the log-sum-exp trick is given as

\begin{equation}
	\label{eq:bhh:selection_mechanism:numerical_stability:log_sum_exp}
	\begin{split}
		LSE(P(h_{k} \vert e_{j}, c_{1};  \boldsymbol{\theta}, \boldsymbol{\phi}, \boldsymbol{\psi})) = \ln(\exp(\phi_{j,k}) +  \exp(\psi_{k}) + \exp(\theta_{k}))
	\end{split}
\end{equation}

\bibliographystyle{elsarticle-num-names}
\bibliography{references.bib}

\end{document}